\documentclass{article}

 \usepackage[dblblindworkshop, final]{neurips_2025}

\usepackage[utf8]{inputenc} 
\usepackage[T1]{fontenc}    
\usepackage{hyperref}       
\usepackage{url}            
\usepackage{booktabs}       
\usepackage{amsfonts}       
\usepackage{nicefrac}       
\usepackage{microtype}      
\usepackage{xcolor}         
\usepackage{graphicx}
\usepackage{subcaption}
\usepackage{amsmath} 

\title{Video Killed the Energy Budget: \\ Characterizing the Latency and Power Regimes of Open Text-to-Video Models}
\workshoptitle{What Makes a Good Video: Next Practices in Video Generation and Evaluation}

\author{
  Julien Delavande \\
  Hugging Face \\
  ENS Paris-Saclay \\
  \texttt{julien.delavande@ens-paris-saclay.fr} \\
  \And
  Regis Pierrard \\
  Hugging Face \\
  \texttt{regis.pierrard@huggingface.co} \\
  \And
  Sasha Luccioni \\
  Hugging Face \\
  \texttt{sasha.luccioni@huggingface.co} \\
}

\begin{document}

\maketitle

\begin{abstract}
Recent advances in text-to-video (T2V) generation have enabled the creation of high-fidelity, temporally coherent clips from natural language prompts. Yet these systems come with significant computational costs, and their energy demands remain poorly understood. In this paper, we present a systematic study of the latency and energy consumption of state-of-the-art open-source T2V models. We first develop a compute-bound analytical model that predicts scaling laws with respect to spatial resolution, temporal length, and denoising steps. We then validate these predictions through fine-grained experiments on WAN2.1-T2V, showing quadratic growth with spatial and temporal dimensions, and linear scaling with the number of denoising steps. Finally, we extend our analysis to six diverse T2V models, comparing their runtime and energy profiles under default settings. Our results provide both a benchmark reference and practical insights for designing and deploying more sustainable generative video systems.
\end{abstract}

\section{Introduction}

Text-to-video (T2V) generation has rapidly become one of the most compelling frontiers of generative AI. Proprietary systems such as OpenAI's \textit{Sora} \citep{videoworldsimulators2024} and DeepMind's \textit{Veo} \citep{deepmind2025veo} have showcased remarkable progress in realism and temporal consistency. At the same time, the open-source community is closing the gap, releasing increasingly powerful models \citep{guo2024animatediffanimatepersonalizedtexttoimage, yang2025cogvideoxtexttovideodiffusionmodels, hacohen2024ltxvideorealtimevideolatent, genmo2024mochi, wan2025wanopenadvancedlargescale} that can be executed on commodity GPUs. As these systems transition from research prototypes to real-world applications used in creative tools and production-grade video synthesis APIs, it becomes crucial to understand not only their quality, but also their computational costs and environmental impacts.

Generating even a few seconds of coherent video typically requires dozens of denoising steps, high spatial resolutions, and hundreds of frames. This leads to substantial energy consumption and long inference times. Yet, most evaluations of T2V models emphasize perceptual metrics such as sample fidelity, FID scores, or motion smoothness, while largely overlooking latency and energy efficiency. In an era where democratization and sustainability are key, these overlooked dimensions deserve systematic study.

In this paper, we make the following contributions:
\begin{itemize}
    \item \textbf{Theoretical analysis.} We develop a compute-bound analytical model of latency and energy for WAN2.1-T2V \citep{wan2025wanopenadvancedlargescale}, decomposing FLOPs by operator and predicting scaling laws with respect to spatial resolution, temporal length, and denoising steps.
    \item \textbf{Empirical validation.} We perform fine-grained microbenchmarks on WAN2.1-T2V to test these predictions, revealing quadratic scaling in spatial and temporal dimensions, and linear scaling in steps.
    \item \textbf{Cross-model benchmarking.} We extend our analysis to six open-source T2V models, comparing their latency and energy profiles under default generation settings.
    \item \textbf{Implications.} We discuss the consequences of these findings for efficient deployment, sustainable model design, and future directions such as diffusion caching and quantization.
\end{itemize}

Together, these contributions provide both a modeling framework and empirical evidence for understanding the structural inefficiencies of T2V pipelines, offering actionable insights for balancing quality and sustainability in generative video systems.
\section{Related Work} 

The environmental costs of machine learning are a new but growing field of scholarship, starting with the pioneering study of Strubell et al., which was the first to quantify the carbon footprint of training a large language model (LLM)~\citeyearpar{Strubell2019}. The subsequent years were marked by more work on the carbon footprint of different types of machine learning (ML) models and the factors that influence them~\cite{Patterson2022,luccioni2022estimatingcarbonfootprintbloom, gupta2021chasing,wu2022sustainable}. While much of the initial work was focused on ML model training -- given that it presents a larger up-front cost in terms of energy and carbon -- recent work has increasingly focused on inference, given the ubiquity of deploying different kinds of ML models in practice. Notably, Luccioni et al.~\citeyearpar{Luccioni2024Facct} carried out the first large-scale study on the energy and carbon costs for different tasks and approches, including image generation.

While there is limited existing work on the energy demands of video generation, recent work by Li et al.~\citeyearpar{li2024carbon}, studied the energy needed to generate videos by the Open-Sora model~\cite{zheng2024open}. They analyzed the energy required to generate 2-second videos at 240p resolution, and found that not only is video generation significantly more energy-intensive than text generation (which corroborates the findings of Luccioni et al.~\citeyearpar{Luccioni2024Facct}), but also that \textit{"the primary source of emissions stemming from iterative diffusion denoising"}. They also found that the energy requirements of video generation scales near-quadratically with video resolution. This is the only existing published study on the energy requirements of video-generation, which is nonetheless limited to a single model and type of output (i.e. video length and resolution), emphasizing the importance of having a better understanding of this important topic. This was the motivation for our own study, which we describe in the following section.

\section{Theoretical Model of Latency and Energy}
\label{sec:theoretical-model}

\begin{figure}[h]
\centering
\includegraphics[width=\linewidth]{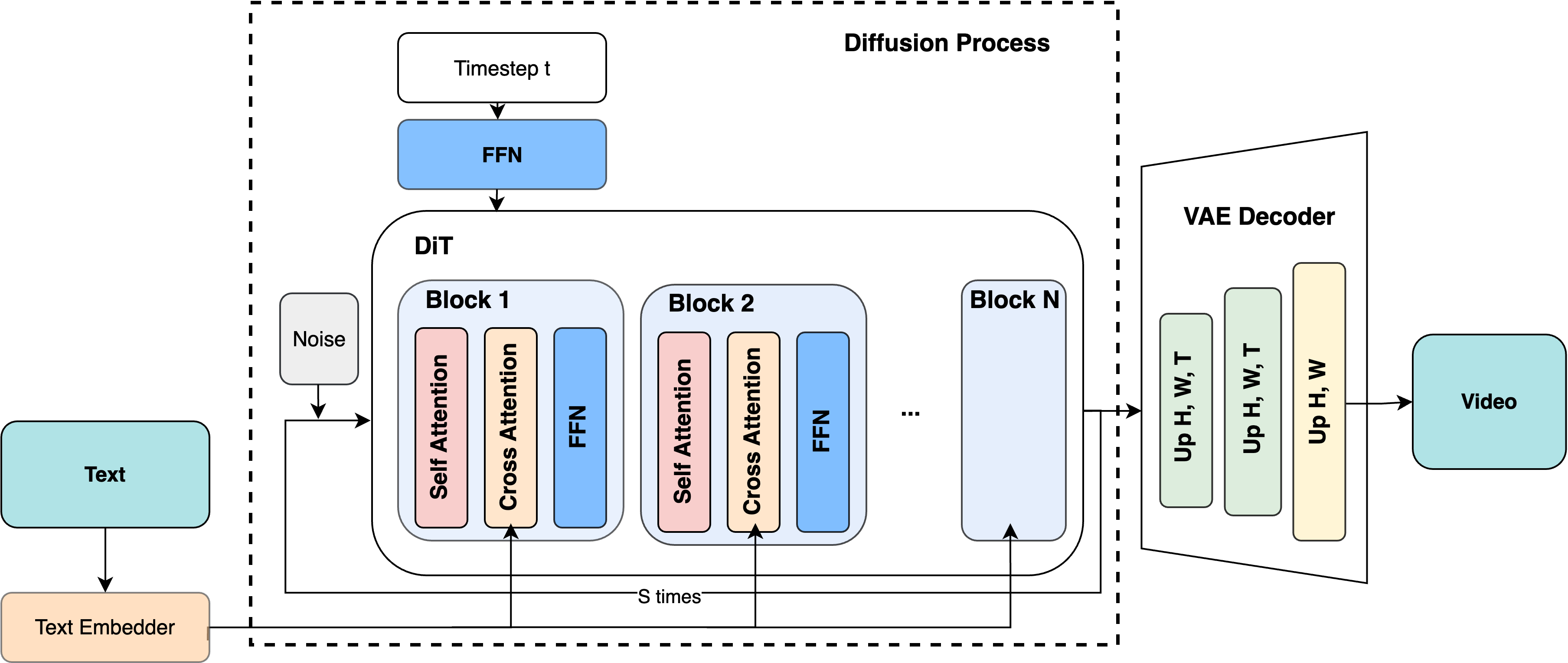}
\caption{Simplified architecture of WAN2.1-T2V-1.3B.}
\label{fig:wan_archi}
\end{figure}

To ground our analysis, we focus on the \textbf{WAN2.1-T2V-1.3B} model
\citep{wan2025wanopenadvancedlargescale}, which serves as our reference architecture. 
WAN2.1 is representative of modern latent text-to-video diffusion systems: 
a pretrained text encoder provides conditioning, a timestep embedding MLP injects
the diffusion step index, a large DiT (Diffusion Transformer) performs the bulk of
spatio-temporal denoising, and a VAE decoder maps latent tensors back to pixel space.
This structure is shown in Figure~\ref{fig:wan_archi}. 
The same framework can be applied to other recent models with minor adjustments. WAN2.1 is also the most downloaded text-to-video model on the Hugging Face Hub at the time of writing, motivating its selection for an in-depth study.

We are then able to derive a compute-bound analytical model of WAN2.1 inference,
decomposing FLOPs by operator and predicting latency and energy as explicit functions
of resolution $(H,W)$, number of frames $T$, and denoising steps $S$.

\subsection{Compute vs. Memory-Bound Regimes}

On modern GPUs such as the NVIDIA H100, inference kernels can be either:
\begin{itemize}
    \item \textbf{Compute-bound}, when execution is limited by arithmetic throughput (FLOP/s).
    \item \textbf{Memory-bound}, when limited by memory bandwidth.
\end{itemize}

Profiling shows that the main operators of WAN2.1 inference
(self-attention, cross-attention, MLPs, VAE convolutions) are predominantly compute-bound.
GPU utilization remains saturated, and power traces indicate negligible CPU-induced idle time.
We therefore adopt a compute-bound model, following the classic roofline formulation
\citep{williams2009roofline}, where latency is proportional to total FLOPs divided by sustained
throughput. This approximation is consistent with prior studies of large-scale transformer
workloads \citep{shoeybi2019megatronlm,narayanan2021efficienttraining,hagemann2024efficientparallelizationlayoutslargescale,jiang2024megascalescalinglargelanguage,pavani2025modelingtemporaldependencesequence}.

\subsection{Notation and Constants}

We follow the HPC convention where one multiply-add corresponds to two FLOPs.  
Throughout, $H\times W$ denotes the spatial resolution, $T$ the number of frames,
$S$ the number of denoising steps, $N$ the number of DiT layers, $d$ the hidden size,
$f$ the MLP expansion factor, $m$ the text conditioning length, $g$ the number of
classifier-free guidance (CFG) passes, and $\ell$ the latent token length seen by the DiT.  
A complete list of symbols, constants, and hardware parameters is provided in
Appendix~\ref{app:flops-derivations}.

The DiT token length $\ell$ grows with the spatial ($H,W$) and temporal ($T$) dimensions of the latent grid:
\[
\ell = \Big(1+\frac{T}{4}\Big)\frac{H}{16}\frac{W}{16}.
\]

\subsection{Operation-Level FLOP Breakdown}

The total FLOPs per video generation can be decomposed into contributions from the text encoder, timestep MLP, the diffusion transformer (DiT), and the VAE decoder, see table~\ref{tab:wan2.1_flops}.  
A full derivation of these FLOP formulas is provided in Appendix~\ref{app:flops-derivations}, where we detail each operator (self-attention, cross-attention, MLP, VAE, text encoder, timestep MLP).

\begin{table}[h]
\centering
\caption{FLOP cost of WAN2.1-T2V-1.3B components. 
Top: once per video. Bottom: per denoising step (to be multiplied by $gS$).
Symbols are defined inline in Section~\ref{sec:theoretical-model}, with the
complete list deferred to Appendix~\ref{app:flops-derivations}.}
\renewcommand{\arraystretch}{1.2}
\begin{tabular}{l l l}
\toprule
\textbf{Component} & \textbf{FLOPs} & \textbf{Notation} \\
\midrule
\multicolumn{3}{c}{\emph{Once per video}} \\
\midrule
Text encoder (T5) &
$\displaystyle 
p_{\text{text}} \, L_{\text{text}} 
\Big( 8 m d_{\text{text}}^2 + 4 m^2 d_{\text{text}} + 4 f_{\text{text}} m d_{\text{text}}^2 \Big)$
& $F_{\text{text}}$ \\[0.2em]

VAE decoder convolutions &
$\displaystyle 
\sum_{j=1}^{N_{\text{dec,conv}}}
2 \, k_t^{(j)} k_h^{(j)} k_w^{(j)} \,
C_{\text{in}}^{(j)} C_{\text{out}}^{(j)} \,
T^{(j)} H^{(j)} W^{(j)}$
& $F_{\text{VAE,conv}}$ \\

VAE decoder 2D “middle” attention &
$\displaystyle 
T_\ast \Big( 8\,C_\ast^2 L_\ast + 4\,L_\ast^2 C_\ast \Big)$
& $F_{\text{VAE,mid-attn}}$ \\
\midrule
\multicolumn{3}{c}{\emph{Per denoising step (multiply by $gS$)}} \\
\midrule
\multicolumn{3}{l}{\textit{DiT}} \\
Self-attention (N layers) &
$\displaystyle N \,\big( 8 \ell d^2 + 4 \ell^2 d \big)$
& $F_{\text{self}}$ \\
Cross-attention (N layers)&
$\displaystyle N \,\big( 4 \ell d^2 + 4 m d^2 + 4 \ell m d \big)$
& $F_{\text{cross}}$ \\
MLP (N layers)&
$\displaystyle N \,\big( 4 f \ell d^2 \big)$
& $F_{\text{mlp}}$ \\
\addlinespace[0.25em]
Timestep MLP (shared across layers) &
$\displaystyle 2\,d_{\tau}\,d \;+\; 14\,d^2$
& $F_{\tau}$ \\
\bottomrule
\end{tabular}
\label{tab:wan2.1_flops}
\end{table}

\subsection{Total FLOPs}

The total FLOPs for generating a video of spatial size $H\times W$, $T$ frames, and $S$ steps is:
\[
F_{\text{total}} = F_{\text{text}} + F_{\text{VAE,conv}}+ 
F_{\text{VAE,mid-attn}} +  
Sg \cdot \big( F_{\text{self}} + F_{\text{cross}} + F_{\text{mlp}} + F_{\tau}\big).
\]
We define $\mu$ as the ratio between sustained and peak throughput:
\[
\mu = \frac{F_{\text{total}}/D_{\text{measured}}}{\Theta_{\text{peak}}}.
\]

Assuming compute-bound execution with empirical efficiency $\mu$, 
and letting $\Theta_{\text{peak}}$ denote the GPU’s theoretical peak throughput in dense BF16,
the total latency $D_{\text{total}}$ of generating a video can be approximated as:
\[
D_{\text{total}} \approx \frac{F_{\text{total}}}{\mu \,\Theta_{\text{peak}}}.
\]

In practice, the H100 provides a dense BF16 peak of $\Theta_{\text{peak}} = 989\,\text{TFLOP/s}$ (\href{https://www.megware.com/fileadmin/user_upload/LandingPage\%20NVIDIA/nvidia-h100-datasheet.pdf}{NVIDIA datasheet}), but this level is unattainable. The empirical efficiency $\mu$ thus acts as a correction factor, reflecting both hardware under-utilization (tile misalignment, kernel overheads, memory-bound ops) and approximations of our latency model. For WAN2.1 -- after performing the experiments explained in section \ref{sec:methodo} -- we obtain $\mu \approx 0.456$, consistent with sustained FLOP utilization of $30$--$63\%$ reported for large-scale transformer inference on H100s \citep{hagemann2024efficientparallelizationlayoutslargescale,jiang2024megascalescalinglargelanguage,pavani2025modelingtemporaldependencesequence}. 
We calibrated $\mu$ by linear regression of measured latencies against theoretical FLOPs across our experiments, which yielded $\mu=0.456$ with negligible overhead and $R^2=0.998$.

\noindent\textbf{Compute-bound regime.}
On the H100, main operators such as self-attention and MLPs become compute-bound above sequence lengths of $\ell \approx 295$ and $\ell \approx 590$, respectively. Since all configurations studied here operate at much higher token counts ($\ell$ is typically in the $10^4$-$10^5$ range even for moderate resolutions such as $480\times 720$ and a few seconds of video), these blocks are firmly compute-bound. For very short $\ell$, the MLP dominates latency and energy, but such regimes are far below our operating range. Full derivation and extensions to other hardware showing the same trends are given in Appendix~\ref{sec:compute_bound_threshold}.

\subsection{Energy Model}

Since sustained GPU power remains close to $P_{\max}$ during inference, the total energy consumed $E_{\text{total}}$:
\[
E_{\text{total}} \approx P_{\max} \cdot D_{\text{total}}.
\]
where $P_{\max}$ denotes the GPU's maximum power draw (here $\sim 700\,\text{W}$).
Thus, energy and latency scale proportionally.

\subsection{Predicted Scaling Regimes}

From these equations, we can anticipate distinct computational regimes:

\begin{itemize}
    \item \textbf{Quadratic scaling in spatial and temporal dimensions.} 
    Since the DiT token length $\ell$ grows linearly with $H$, $W$, and $T$, 
    the self- and cross-attention terms contribute $\mathcal{O}(\ell^2)$ FLOPs, 
    leading to quadratic growth in latency and energy as resolution or frame count increases.
    \item \textbf{Linear scaling in denoising steps.} 
    Each step applies the same sequence of $N$ transformer layers, so the 
    ideal cost scales as $\mathcal{O}(S)$.
    \item \textbf{Negligible contributions from auxiliary components.} 
    The text encoder is run once per video, and the timestep MLP adds 
    only a small overhead per step. Likewise, the VAE decoder scales linearly 
    with voxel count $T\!\times\!H\!\times\!W$ and is quickly dominated by 
    the quadratic DiT cost.
\end{itemize}

In summary, the theoretical model predicts that WAN2.1 inference is 
\emph{transformer-dominated and compute-bound}, with quadratic regimes in spatial 
and temporal dimensions, linear dependence on denoising steps, and minor overhead 
from conditioning networks. These predictions will be validated against empirical 
measurements in Section~\ref{sec:empirical_findings}.

\section{Methodology}
\label{sec:methodo}

Our methodology combines two complementary perspectives.  
First, we perform controlled micro-benchmarks on \textbf{WAN2.1-T2V-1.3B}, our reference model,
to validate the scaling regimes predicted by the theoretical model (Section~\ref{sec:theoretical-model}).
Second, we benchmark a diverse set of recent open-source text-to-video models under default settings,
to situate WAN2.1 within the broader ecosystem.

\subsection{Hardware and Measurement Protocol}

All experiments were conducted on a dedicated NVIDIA H100 SXM GPU (80GB HBM3) 
paired with an 8-core AMD EPYC 7R13 CPU, with no co-scheduled jobs.  
We measured GPU and CPU energy using \texttt{CodeCarbon}~\citep{codecarbon}, 
which interfaces with NVML and pyRAPL, and estimated RAM energy using CodeCarbon’s default heuristic\footnote{\url{https://mlco2.github.io/codecarbon/methodology.html\#ram}}.  

To reduce noise, each measurement included two warmup iterations, followed by 
five repeated runs. Inference used the Hugging Face 
\texttt{Diffusers} library~\cite{von-platen-etal-2022-diffusers} with default generation parameters. We relied on the standard optimizations provided by recent PyTorch releases, such as fused kernels and FlashAttention~\citep{dao2023flashattention}, which are automatically enabled.

\subsection{Controlled Scaling Experiments on WAN2.1-T2V-1.3B}

To validate the theoretical model, we systematically varied the three key structural parameters:
resolution, number of frames, and denoising steps. 
Since the text encoder always pads or truncates prompts to a fixed length of 512 tokens, 
the specific choice of prompt does not affect runtime. 
We therefore fixed a single prompt and applied the same warmup-and-repetition protocol as above 
to isolate structural scaling laws.

\begin{itemize}
  \item \textbf{Spatial resolution:} from 256$\times$256 to 3520$\times$1980, 
  both dimensions divisible by 8 (model constraint). Frames and steps fixed.
  \item \textbf{Temporal length (frames):} from 4 to 100 in increments of 4 (model constraint). Resolution and steps fixed.
  \item \textbf{Denoising steps:} from 1 to 200. Resolution and frames fixed.
\end{itemize}

For each configuration we logged total latency (seconds) and energy for each hardware component (GPU / CPU / RAM).

\subsection{Cross-Model Benchmark}

To provide a bird’s-eye view of energy and latency costs across current systems,
we selected a diverse set of models spanning different architectures and parameter scales
(Table~\ref{tab:models}), focusing on those that are among the most downloaded and trending on the Hugging Face Hub at the time of writing.

For this benchmark, we generated \textbf{50 different prompts} per model. 
Each prompt was measured with the protocol above (2 warmups, 5 runs), yielding robust averages 
and standard deviations that capture both runtime noise and input variability.

\begin{itemize}
  \item \textbf{AnimateDiff}~\citep{guo2024animatediffanimatepersonalizedtexttoimage}(\href{https://huggingface.co/spaces/CompVis/stable-diffusion-license}{License}
) - lightweight motion-layer diffusion.
  \item \textbf{CogVideoX-2b/5b}~\citep{yang2025cogvideoxtexttovideodiffusionmodels} (\href{https://huggingface.co/zai-org/CogVideoX-5b/blob/main/LICENSE}{License}) - cascaded base + refiner stages.
  \item \textbf{LTX-Video-0.9.7-dev}~\citep{hacohen2024ltxvideorealtimevideolatent}(\href{https://huggingface.co/Lightricks/LTX-Video/blob/main/LTX-Video-Open-Weights-License-0.X.txt}{License}) - autoregressive temporal modeling.
  \item \textbf{Mochi-1-preview}~\citep{genmo2024mochi}(\href{https://huggingface.co/datasets/choosealicense/licenses/blob/main/markdown/apache-2.0.md}{License}) - large-scale diffusion optimized for motion realism.
  \item \textbf{WAN2.1-T2V (1.3B and 14B)}~\citep{wan2025wanopenadvancedlargescale}(\href{https://huggingface.co/datasets/choosealicense/licenses/blob/main/markdown/apache-2.0.md}{License}) - high-resolution latent diffusion with DiT backbone.
\end{itemize}

\begin{table}[h]
\centering
\caption{Default generation settings for each model (from Hugging Face model cards).}
\begin{tabular}{lcccc}
\toprule
\textbf{Model} & \textbf{Steps} & \textbf{Resolution (HxW)} & \textbf{Frames} & \textbf{FPS} \\
\hline
AnimateDiff           & 4   & 512$\times$512  & 16   & 10 \\
CogVideoX-2b          & 50  & 480$\times$720  & 49   & 8  \\
CogVideoX-5b          & 50  & 480$\times$720  & 49   & 8  \\
LTX-Video             & 40   & 512$\times$704  & 121  & 24 \\
Mochi-1-preview       & 64  & 480$\times$848  & 84   & 30 \\
WAN2.1-T2V-1.3B       & 50  & 720$\times$1280  & 81   & 15 \\
WAN2.1-T2V-14B        & 50  & 720$\times$1280  & 81   & 15 \\
\bottomrule
\end{tabular}

\label{tab:models}
\end{table}

We did not assess perceptual quality to isolate compute behavior; instead, these experiments confront the predicted quadratic and linear regimes (Section~\ref{sec:theoretical-model}) with actual scaling laws and scheduler-induced deviations. All code, prompts, and configurations are available in an anonymized repository at \href{https://github.com/anonymous-222103/video-killed-energy-budget}{GitHub repo}, and all generated videos are released on the Hugging Face Hub at \href{https://huggingface.co/VideoKilledEnergyBudget}{HF org}.

\section{Empirical Findings}
\label{sec:empirical_findings}

We now compare the theoretical predictions of Section~\ref{sec:theoretical-model} with
empirical measurements -- first by conducting a fine-grained validation on \textbf{WAN2.1-T2V-1.3B} and
comparing measured energy and latency against theoretical curves as resolution, temporal
length, and denoising steps vary. We then situate these results in the broader context
of other open-source video generation models.

\subsection{Validation on WAN2.1-T2V-1.3B}

In this section we focus exclusively on \emph{GPU energy and latency}, since GPU accounts
for 80--90\% of the total consumption and dominates inference cost. Figures show theoretical
predictions (stacked areas by operator: self-attention, cross-attention, MLP, VAE, text encoder,
timestep MLP) with empirical measurements overlaid as points with error bars.

\subsubsection{Spatial Resolution}

\begin{figure}[h]
\centering
\begin{minipage}[b]{0.48\linewidth}
  \centering
  \includegraphics[width=\linewidth]{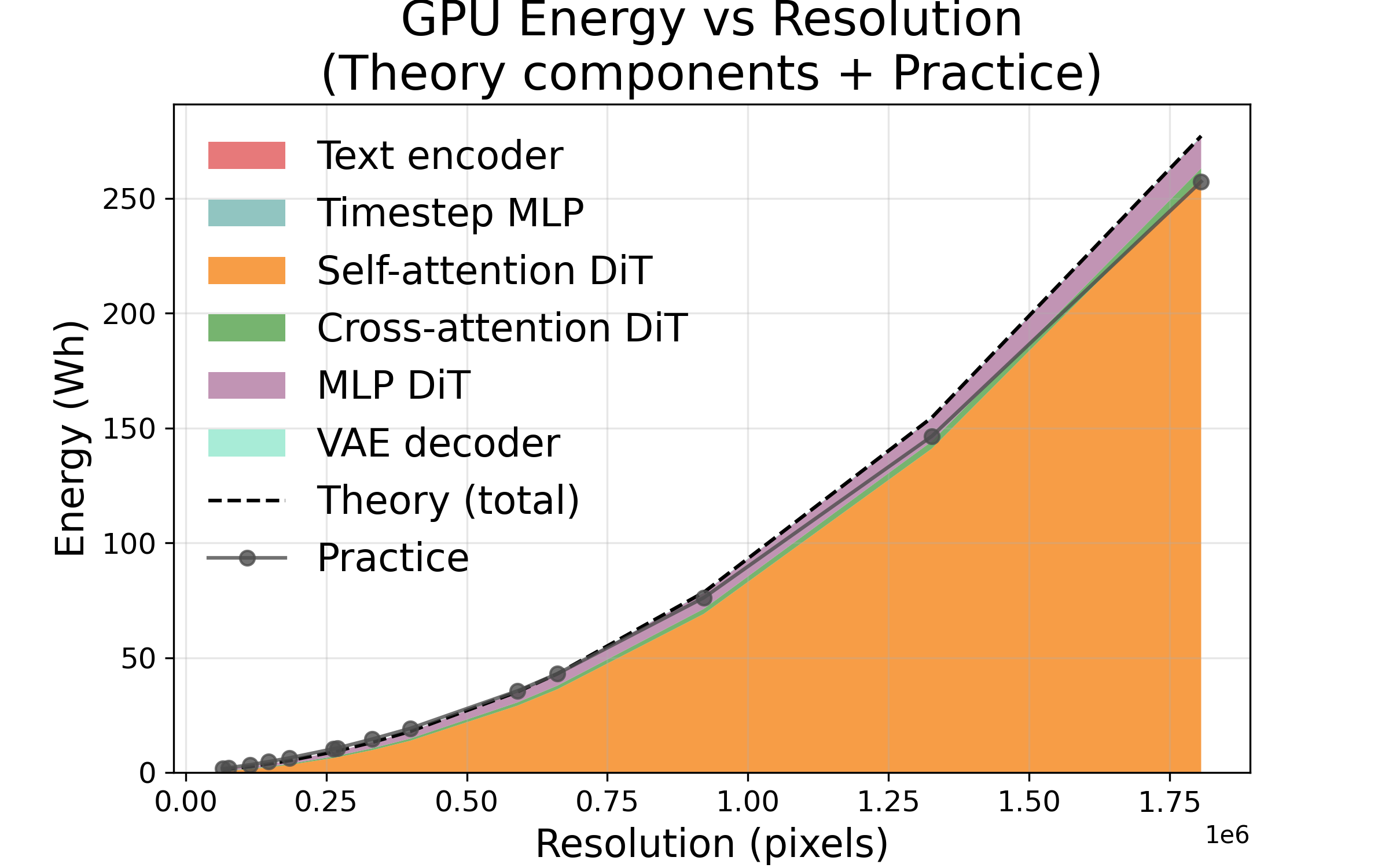}
  \caption*{(a) GPU energy vs. spatial resolution}
\end{minipage}
\hfill
\begin{minipage}[b]{0.48\linewidth}
  \centering
  \includegraphics[width=\linewidth]{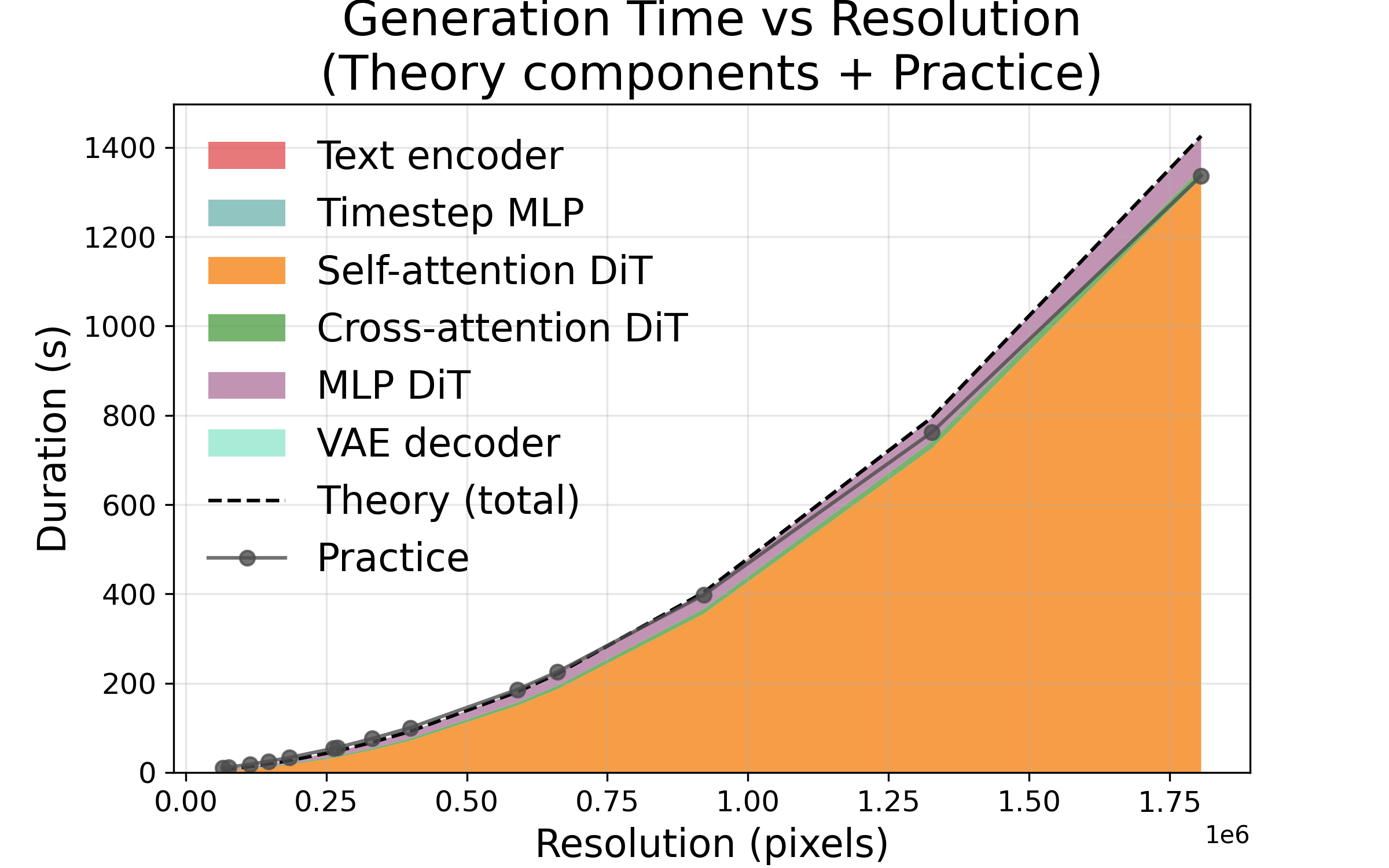}
  \caption*{(b) Latency vs. spatial resolution}
\end{minipage}
\caption{Empirical results (points) vs. theoretical predictions (stacked areas per operator) 
as a function of resolution. Both energy and latency follow the predicted quadratic regime.}
\label{fig:resolution_scaling}
\end{figure}

Increasing the resolution from 256$\times$256 to 3520$\times$1980 (frames - 81 and steps - 50 fixed)
\emph{causes both latency and energy to grow quadratically}. 
Theoretical predictions (stacked by operator) and empirical measurements are compared
in Figure~\ref{fig:resolution_scaling}. 
The agreement remains strong across the entire range, with modest deviations at high resolutions
(see Table~\ref{tab:mpe}). The VAE contribution remains minor compared to the DiT blocks.

\subsubsection{Temporal Length (Frames)}

\begin{figure}[h]
\centering
\begin{minipage}[b]{0.48\linewidth}
  \centering
  \includegraphics[width=\linewidth]{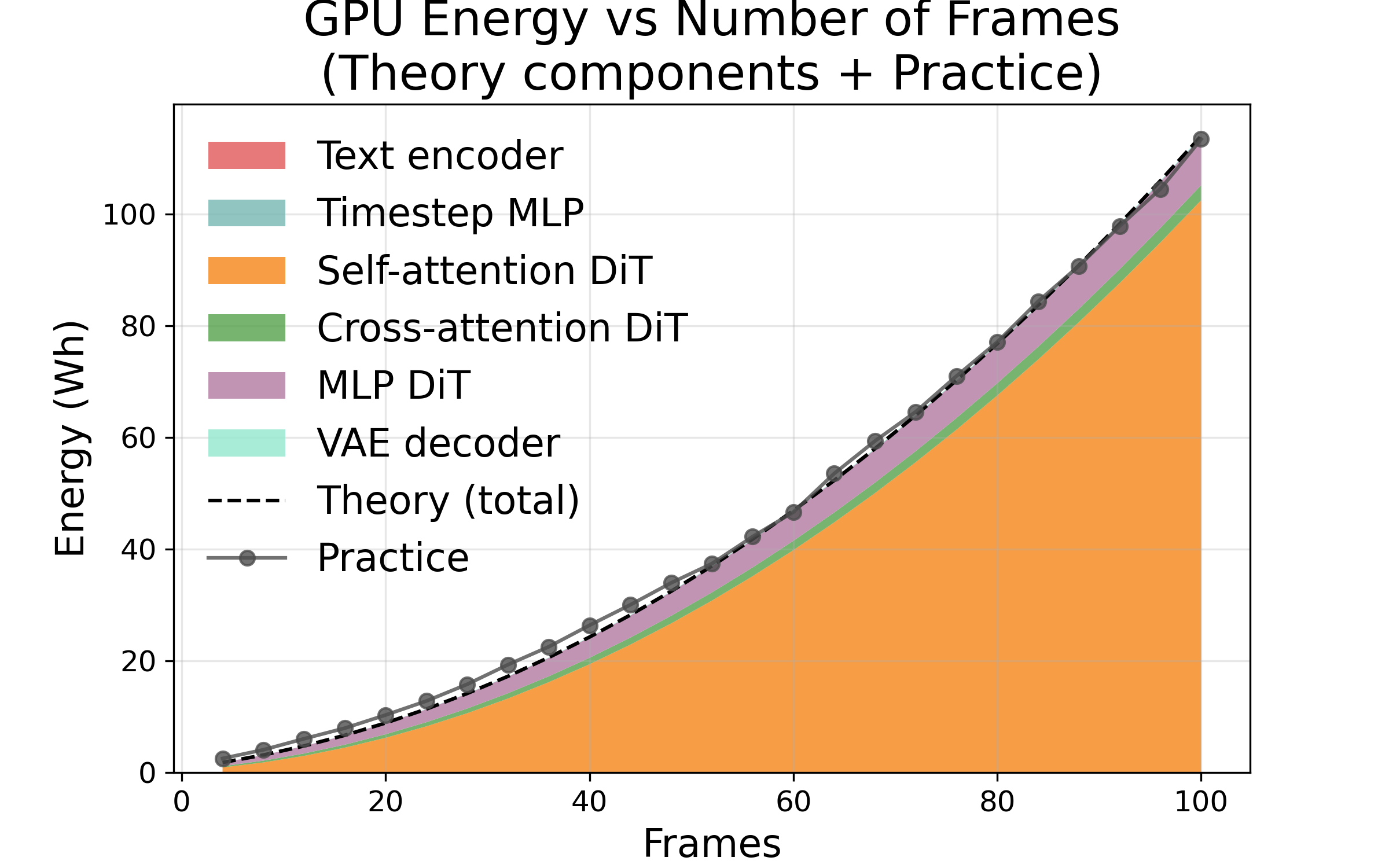}
  \caption*{(a) GPU energy vs. number of frames}
\end{minipage}
\hfill
\begin{minipage}[b]{0.48\linewidth}
  \centering
  \includegraphics[width=\linewidth]{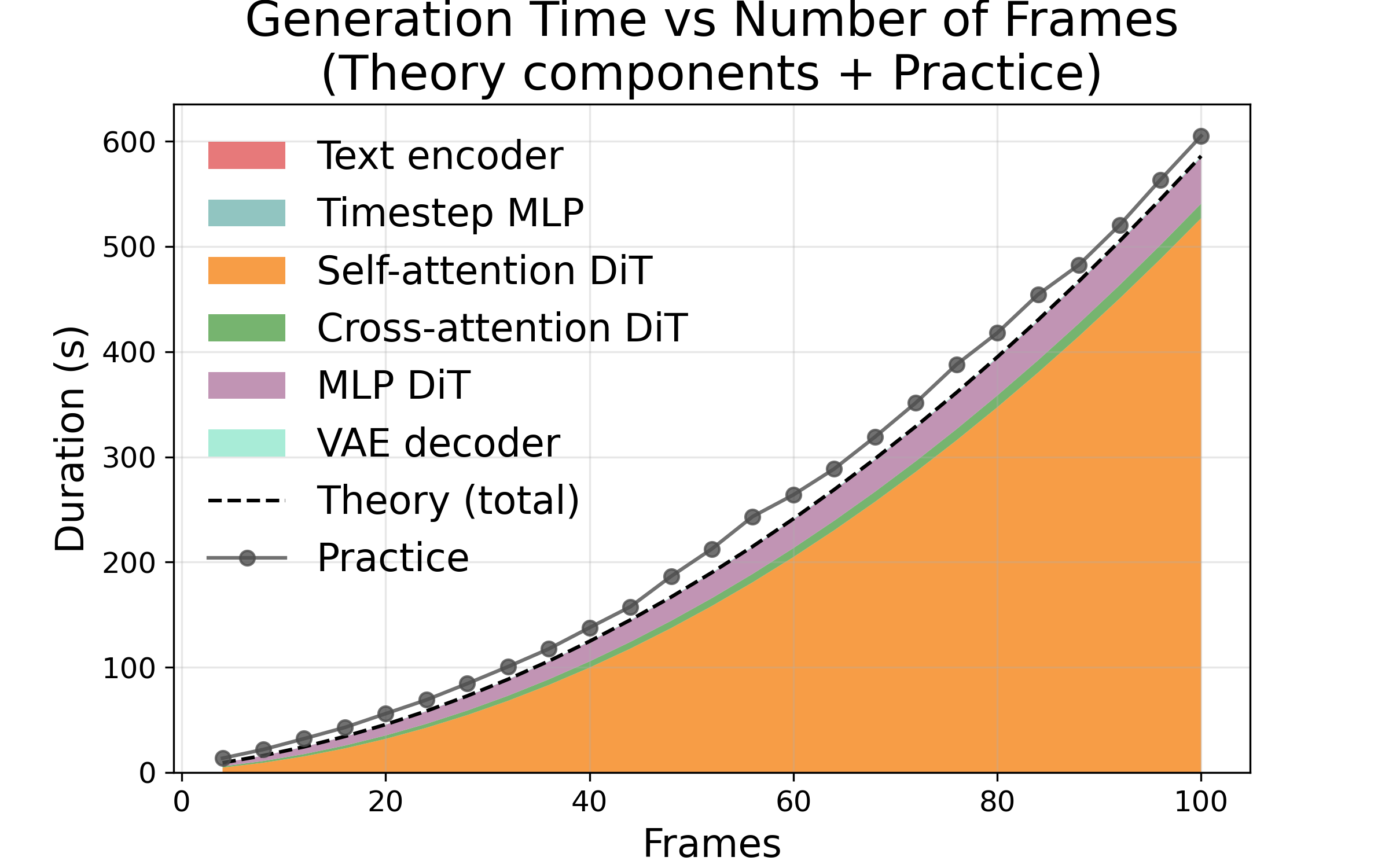}
  \caption*{(b) Latency vs. number of frames}
\end{minipage}
\caption{Empirical results (points) vs. theoretical predictions (stacked areas per operator) 
as a function of temporal length. Both metrics follow the quadratic regime predicted by the model.}
\label{fig:frame_scaling}
\end{figure}

Varying the number of frames from 4 to 100 (resolution - $720\times1280$ and steps - 50 fixed) 
also induces \emph{quadratic growth} in both latency and energy, as shown in
Figure~\ref{fig:frame_scaling}. 
This behavior directly follows from the quadratic dependence of attention on the token count $\ell$.
The model closely tracks empirical results, with errors reported in Table~\ref{tab:mpe}.

\subsubsection{Denoising Steps}

\begin{figure}[h]
\centering
\begin{minipage}[b]{0.48\linewidth}
  \centering
  \includegraphics[width=\linewidth]{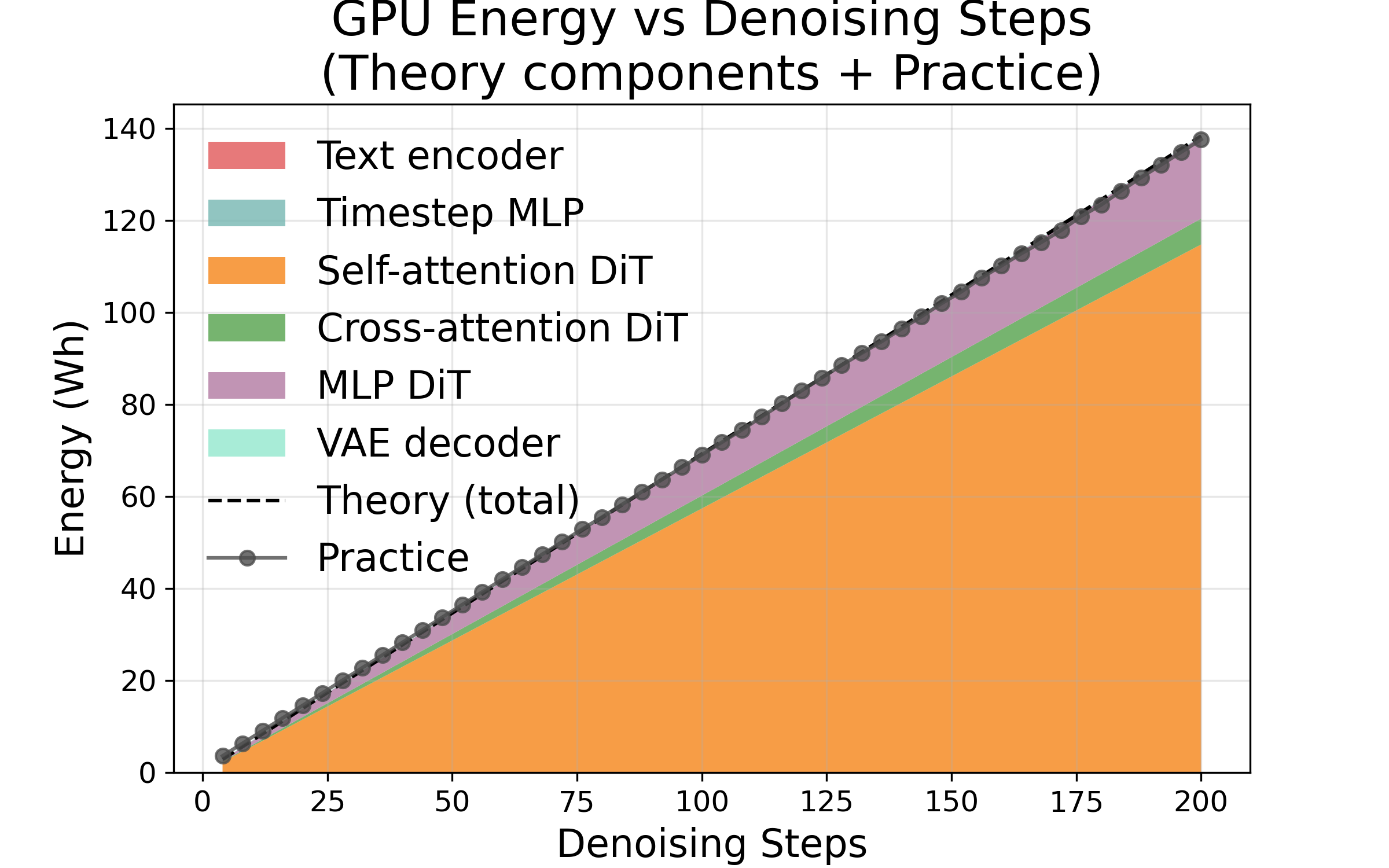}
  \caption*{(a) GPU energy vs. denoising steps}
\end{minipage}
\hfill
\begin{minipage}[b]{0.48\linewidth}
  \centering
  \includegraphics[width=\linewidth]{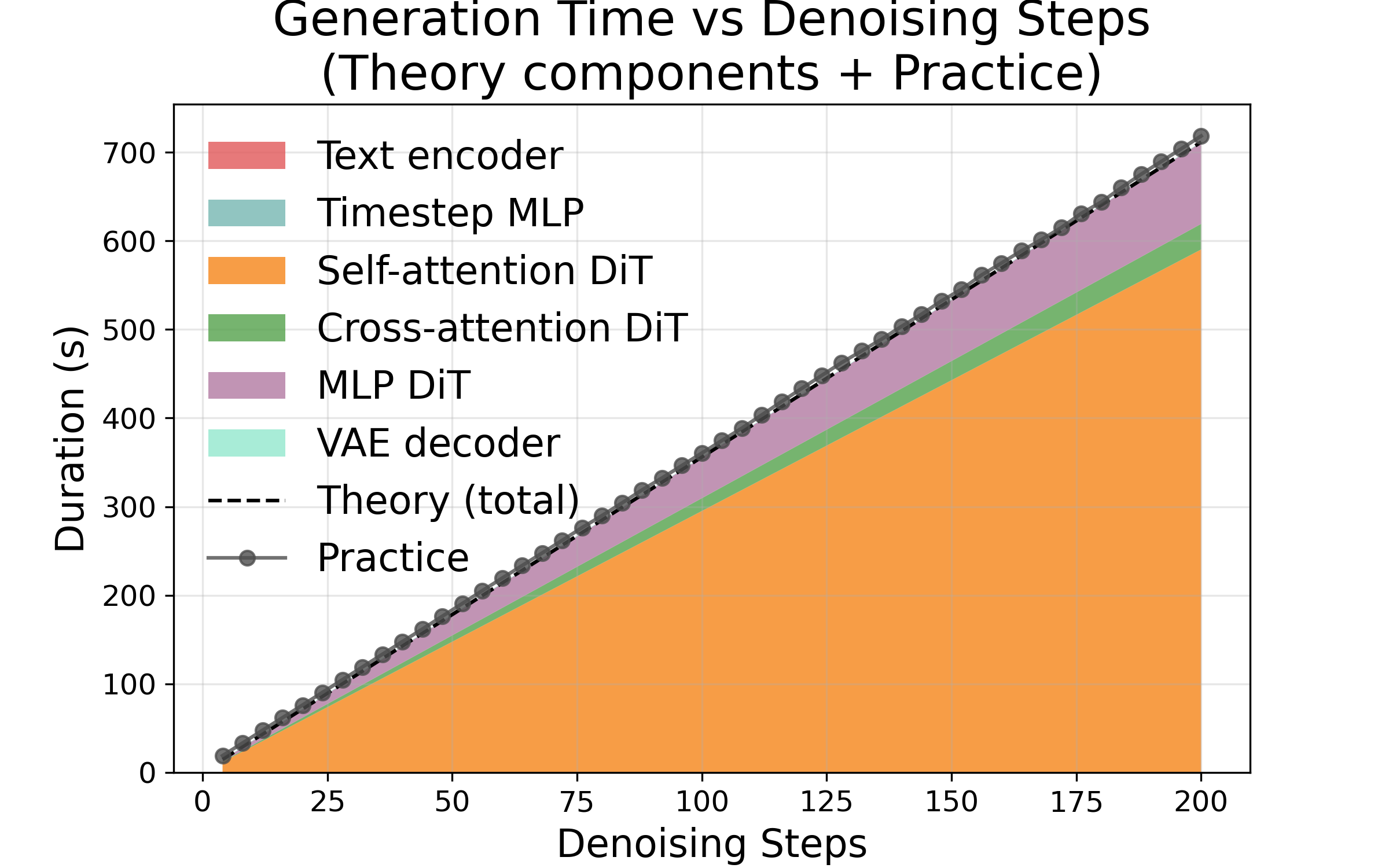}
  \caption*{(b) Latency vs. denoising steps}
\end{minipage}
\caption{Empirical results (points) vs. theoretical predictions (stacked areas per operator) 
as a function of denoising steps. Both energy and latency scale linearly with $S$, in near-perfect 
agreement with the compute-bound model.}
\label{fig:steps_scaling}
\end{figure}

In contrast to resolution and frame count (resolution - $720\times1280$ and frames - 81 fixed), scaling with the number of denoising steps 
is \emph{perfectly linear}, exactly as predicted by the theoretical model. 
Each additional step applies the same $N$ transformer layers, leading to a cost that grows 
proportionally with $S$. Figure~\ref{fig:steps_scaling} shows near-perfect alignment 
between predictions and measurements, with errors below 2\% (Table~\ref{tab:mpe}).

\begin{table}[h]
\centering
\caption{Mean percentage error (MPE) between theoretical predictions and empirical measurements.}
\begin{tabular}{lcc}
\toprule
 & \textbf{Energy} & \textbf{Latency} \\
\midrule
Resolution scaling   & 11.6\% & 14.0\% \\
Temporal length      & 6.6\%  & 10.5\% \\
Denoising steps      & 1.9\%  & 1.9\%  \\
\bottomrule
\end{tabular}
\label{tab:mpe}
\end{table}

\subsection{Cross-Model Comparison}

\begin{figure}[h]
\centering
\begin{minipage}[b]{0.48\linewidth}
  \centering
  \includegraphics[width=\linewidth]{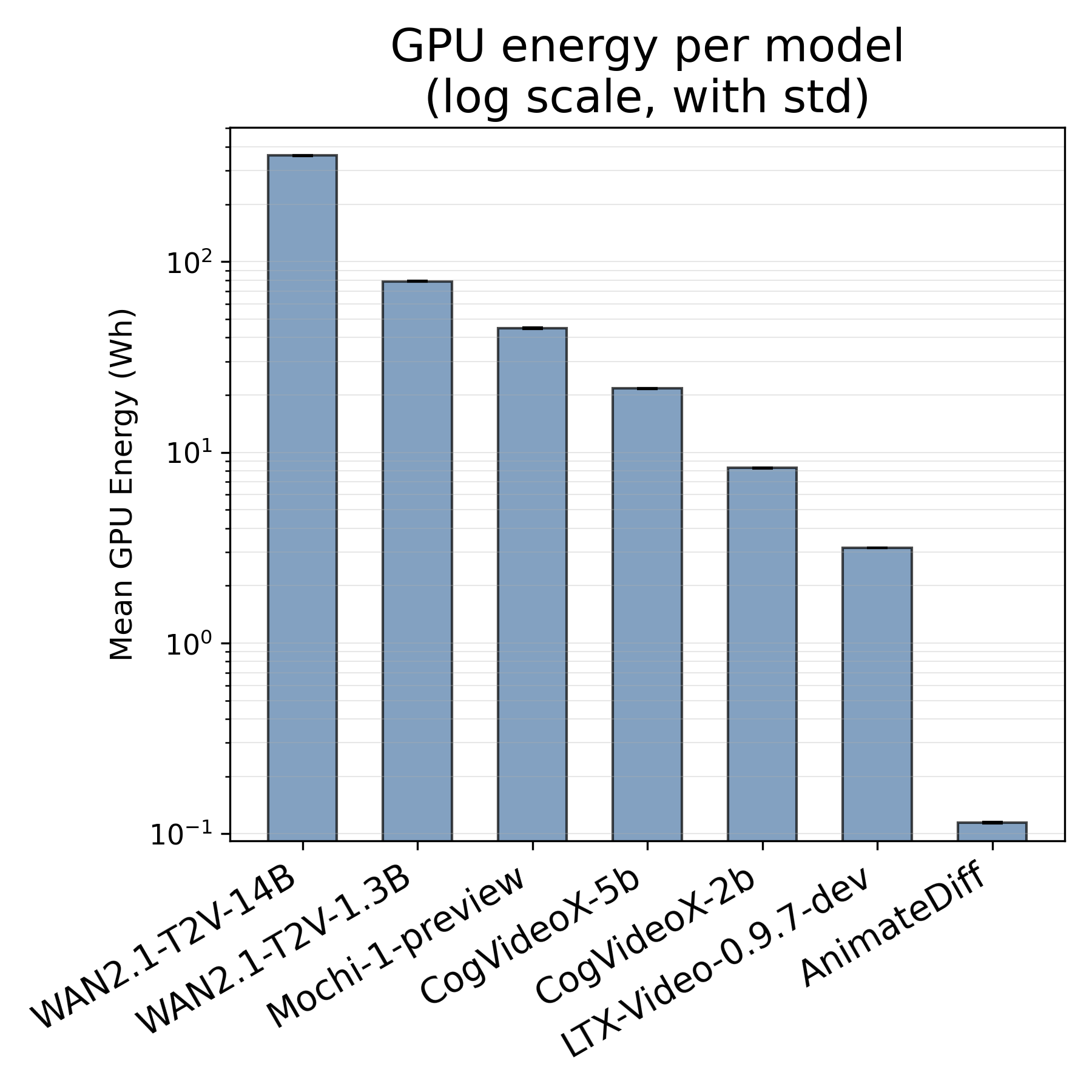}
  \caption*{(a) GPU energy per model for one video}
\end{minipage}\hfill
\begin{minipage}[b]{0.48\linewidth}
  \centering
  \includegraphics[width=\linewidth]{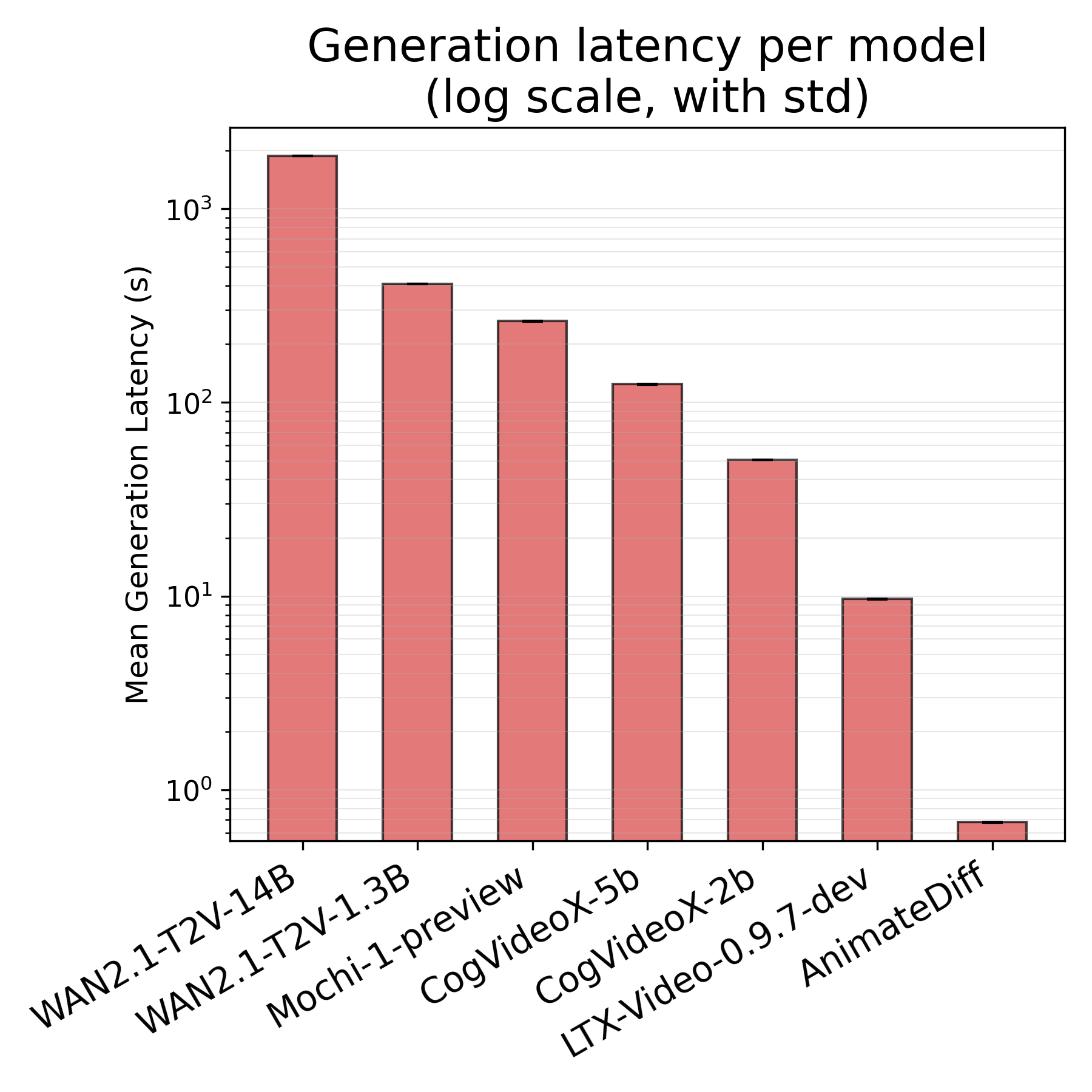}
  \caption*{(b) Generation latency per model for one video}
\end{minipage}

\vspace{0.5cm}

\begin{minipage}[b]{\linewidth}
  \centering
  \includegraphics[width=\linewidth]{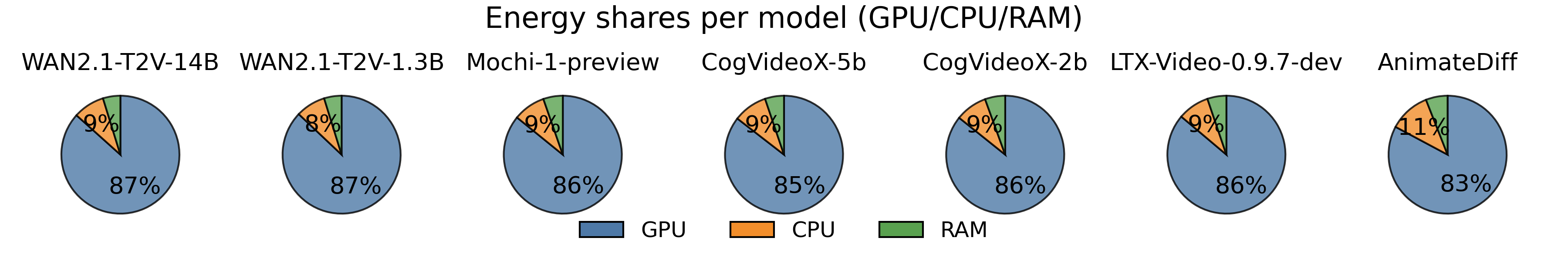}
  \caption*{(c) Energy shares (GPU/CPU/RAM)}
\end{minipage}

\caption{Cross-model comparison of energy and latency. 
Top: GPU energy and latency (log scale, with std). 
Bottom: relative contributions of GPU, CPU, and RAM.}
\label{fig:cross_model_comparison}
\end{figure}

Finally, we compare average GPU energy consumption, latency, and component-wise
energy shares across seven open-source text-to-video models under their default
generation settings (Figure~\ref{fig:cross_model_comparison}).

\begin{table}[h]
\centering
\caption{Cross-model average latency and energy consumption (default settings). 
All values are reported as mean $\pm$ std.}
\label{tab:cross_model_summary}
\renewcommand{\arraystretch}{1.15}
\begin{tabular}{lcccc}
\toprule
\textbf{Model} & \textbf{Latency (s)} & \textbf{GPU (Wh)} & \textbf{CPU (Wh)} & \textbf{RAM (Wh)} \\
\midrule
WAN2.1-T2V-14B      & $1875 \pm 2.1$   & $359.7 \pm 0.5$   & $35.6 \pm 4.0$   & $19.8 \pm 0.02$ \\
WAN2.1-T2V-1.3B     & $410 \pm 0.5$    & $78.8 \pm 0.1$    & $7.4 \pm 0.4$    & $4.3 \pm 0.01$ \\
Mochi-1-preview     & $263 \pm 0.5$    & $44.7 \pm 0.2$    & $4.6 \pm 0.01$   & $2.8 \pm 0.01$ \\
CogVideoX-5B        & $124 \pm 0.4$    & $21.6 \pm 0.05$   & $2.4 \pm 0.03$   & $1.3 \pm 0.004$ \\
CogVideoX-2B        & $50.6 \pm 0.2$   & $8.3 \pm 0.03$    & $0.84 \pm 0.04$  & $0.53 \pm 0.002$ \\
LTX-Video-0.9.7-dev & $9.7 \pm 0.01$   & $3.16 \pm 0.006$  & $0.32 \pm 0.002$ & $0.19 \pm 0.001$ \\
AnimateDiff         & $0.68 \pm 0.002$ & $0.115 \pm 0.001$ & $0.016 \pm 0.0001$ & $0.008 \pm 0.00003$ \\
\bottomrule
\end{tabular}
\end{table}

We observe orders-of-magnitude disparities: \textbf{AnimateDiff} requires only
\textbf{0.14 Wh} in total, while \textbf{WAN2.1-T2V-14B} consumes over
\textbf{415 Wh}, a factor of nearly 3000$\times$. Latency follows a similar trend,
with lightweight models producing clips in less than a second, while large-scale
architectures such as WAN2.1-14B or Mochi require several minutes of inference.
 These differences stem from:
\begin{itemize}
  \item \textbf{Model size:} larger models (WAN2.1-14B, Mochi) process more parameters per step.  
  \item \textbf{Sampling steps:} AnimateDiff runs in 4 steps vs.\ 60--64 for others.  
  \item \textbf{Video length:} frame count and FPS vary significantly.  
  \item \textbf{Architectural complexity:} cascaded pipelines (CogVideoX) require multiple stages.  
\end{itemize}

As shown in the bottom panel, GPU consistently dominates energy consumption
($>$80\%) across all models, confirming a compute-bound regime with high GPU
utilization. CPU and RAM contributions remain secondary, though slightly more
pronounced in cascaded or multi-stage pipelines.

\section{Discussion}

Our results confirm that WAN2.1 inference operates in a \textbf{compute-bound regime}, where latency and energy scale quadratically with spatial $(H,W)$ and temporal $(T)$ dimensions, and linearly with denoising steps $(S)$. The close match between theory and measurement validates the analytical model and provides clear guidance for practitioners.

\paragraph{Implications for efficiency.}
Quadratic scaling in $H$, $W$, and $T$ means that even modest increases in resolution or video length incur steep costs: doubling any of these dimensions in isolation yields $\sim 4\times$ more compute, while scaling multiple dimensions compounds multiplicatively (e.g., $H$ and $W$ doubled $\to 16\times$). Thus, \emph{output size control} is a powerful lever: reducing spatial or temporal length often saves more than architectural changes. In practice, offering presets (e.g., ``low resolution, low frames'' vs.\ ``high fidelity'') balances user needs with energy cost.

\paragraph{Validated linear regime in steps.}
In contrast, denoising steps scale linearly, with measured costs matching theoretical predictions once empirical efficiency $\mu$ is applied. This makes $S$ a reliable knob for latency–quality trade-offs: halving steps roughly halves both latency and energy.

\paragraph{Opportunities for model-level improvements.}
The public Hugging Face implementation of WAN2.1 lacks inference-time optimizations, but the original paper suggests effective techniques: (i) \emph{diffusion caching}, reusing redundant attention/CFG activations for up to $1.62\times$ savings, and (ii) \emph{quantization}, using FP8/INT8 mixed precision for $\sim 1.27\times$ speedup without loss. Other avenues include step pruning, low-rank attention, and kernel fusion to better exploit GPU tensor cores.

\paragraph{Broader implications.}
Video diffusion is far more costly than text or image generation. 
Normalized per output, Luccioni et al.\ \citep{Luccioni2024Facct} report average costs of 
$\sim$0.002 Wh for text classification, 0.047 Wh for text generation, and 2.9 Wh for image generation. 
By comparison, generating a single short video with WAN2.1–T2V–1.3B consumes nearly $\sim$90 Wh. 
This places video diffusion roughly $30\times$ more costly than image generation, 
$2{,}000\times$ than text generation, and $45{,}000\times$ than text classification. 
At scale, the quadratic growth in $(H,W,T)$ implies rapidly increasing hardware and environmental costs, 
highlighting the need for hardware-aware optimizations and sustainable model design. Theoretical thresholds derived in Appendix~\ref{sec:compute_bound_threshold} suggest that compute-bound behavior extends to all tested accelerators, reinforcing the generality of our scaling model.

\section{Limitations and Conclusion}

\paragraph{Limitations.}
Our analysis provides a detailed characterization of WAN2.1–1.3B using the open-source Hugging Face codebase. As such, it does not capture potential improvements from internal optimizations such as diffusion caching, quantization, or kernel fusion. The theoretical model also assumes uniform attention cost and ignores memory hierarchy effects, which may cause deviations for small inputs or extreme aspect ratios. 

Energy measurements were conducted on a single hardware platform (NVIDIA H100 SXM). While Appendix~\ref{sec:compute_bound_threshold} shows that the compute-bound regime and associated scaling trends should extend to other accelerators for realistic token lengths, this remains to be confirmed experimenta. We deliberately excluded perceptual quality from our scope, leaving open the question of energy–fidelity tradeoffs. Finally, many production T2V systems (e.g., Veo) also generate audio, whose contribution to energy cost remains unexplored.

\paragraph{Conclusion.}
We presented a systematic study of latency and energy consumption in text-to-video generation. Through fine-grained experiments on WAN2.1, we validated a simple analytical model that predicts quadratic scaling with spatial and temporal dimensions, and linear scaling with denoising steps. Cross-model benchmarks confirmed that this compute-bound regime extends broadly across recent open-source systems, with orders-of-magnitude disparities in cost depending on model size, sampling strategy, and video length. 

These findings highlight both the structural inefficiency of current video diffusion pipelines and the urgent need for efficiency-oriented design. Promising avenues include diffusion caching, low-precision inference, step pruning, and improved attention mechanisms. We hope this work serves as both a benchmark reference and a modeling framework to guide future research on sustainable generative video systems.

\newpage
\bibliographystyle{plainnat}
\bibliography{main}
\newpage

\appendix
\appendix
\section{Detailed FLOP Derivations and Scaling Laws}
\label{app:flops-derivations}

\paragraph{Conventions.}
We follow the HPC convention where one multiply–add equals two FLOPs.
Matrix multiplications of shape $(a\times b)\cdot(b\times c)$ therefore cost
$2abc$ FLOPs. Bias additions, activations, layer norms, and softmax are lower
order and omitted unless stated. All results below apply per forward pass.

\begin{table}[h!]
\centering
\caption{Complete set of WAN2.1-T2V-1.3B hyperparameters and constants.
This table provides the full notation, including VAE layer-wise symbols
(instantiated explicitly in Appendix~\ref{app:vae}).}

\begin{tabular}{l c l}
\toprule
Symbol & Value & Meaning \\
\midrule
\multicolumn{3}{c}{\emph{Global video parameters}} \\
\midrule
$T$ & variable & Number of frames \\
$H \times W$ & variable & Input spatial resolution \\
$S$ & variable & Number of denoising steps \\
$g$ & 2 & CFG passes per step (cond + uncond) \\
$v_t, v_s$ & $4, 8$ & Temporal and spatial downsampling factors of the VAE \\
$p_h, p_w$ & $2, 2$ & Spatial patch size in the DiT latent grid \\
\midrule
\multicolumn{3}{c}{\emph{Diffusion Transformer (DiT)}} \\
\midrule
$N$ & 32 & Number of DiT layers \\
$d$ & 2048 & Hidden size \\
$f$ & 4 & MLP expansion factor ($8192=4d$) \\
$\ell$ & $(1+\tfrac{T}{4})\tfrac{H}{16}\tfrac{W}{16}$ & Token length of latent grid \\
\midrule
\multicolumn{3}{c}{\emph{Text encoder (T5-XXL)}} \\
\midrule
$m$ & 512 & Output tokens per video (conditioning length) \\
$p_{\text{text}}$ & 2 & Calls per video (cond + uncond) \\
$d_{\text{text}}$ & 4096 & Hidden size \\
$L_{\text{text}}$ & 24 & Encoder layers \\
$f_{\text{text}}$ & 2.5 & MLP expansion factor \\
\midrule
\multicolumn{3}{c}{\emph{Timestep embedding}} \\
\midrule
$d_{\tau}$ & 256 & Hidden width of timestep MLP \\
\midrule
\multicolumn{3}{c}{\emph{VAE (layer-wise; values in App.~\ref{app:vae})}} \\
\midrule
$j$ & $1,\dots,N_{\text{dec,conv}}$ & Layer index along the VAE \textbf{decoder} path \\
$N_{\text{dec,conv}}$ & 11 & Number of 3D conv layers in the VAE decoder \\
$k_t^{(j)},k_h^{(j)},k_w^{(j)}$ & -- & 3D kernel sizes of decoder layer $j$ \\
$C_{\text{in}}^{(j)},C_{\text{out}}^{(j)}$ & -- & In/out channels at decoder layer $j$ \\
$T^{(j)},H^{(j)},W^{(j)}$ & -- & Output grid sizes at decoder layer $j$ \\
$C_\ast$ & 384 & Channel width at middle attention block \\
$T_\ast,H_\ast,W_\ast$ & $\lceil T/4\rceil, H/8, W/8$ & Grid sizes at middle resolution \\
$L_\ast$ & $H_\ast W_\ast$ & Spatial token length per frame (2D middle attention) \\
\midrule
\multicolumn{3}{c}{\emph{Hardware / efficiency constants}} \\
\midrule
$\mu$ & 0.456 & Empirical efficiency (fraction of $\Theta_{\text{peak}}$) \\
$\Theta_{\text{peak}}$ & $989{\times}10^{15}$ FLOP/s & Peak GPU throughput (H100) \\
$P_{\max}$ & 700 W & Sustained GPU power \\
$D_{\text{total}}$ & $F_{\text{total}}/(\mu \Theta_{\text{peak}})$ & Total latency \\
\bottomrule
\end{tabular}
\label{tab:wan2.1_hparams}
\end{table}

\subsection{Latent Tokenization and Shapes}
\label{app:shapes}
Let the video have $T$ frames and spatial size $H\times W$ in pixels.
The VAE downsamples temporally by a factor $v_t$ and spatially by $v_s$,
and the DiT operates on spatial patches of size $p_h\times p_w$ in the latent grid.
The token length $\ell$ seen by the DiT is
\begin{equation}
\label{eq:ell}
\ell \;=\; \Big(1 + \frac{T}{v_t}\Big)\;\frac{H}{v_s\,p_h}\;\frac{W}{v_s\,p_w}\,.
\end{equation}
In WAN2.1 we use $(v_t,v_s,p_h,p_w)=(4,8,2,2)$,
hence the shorthand $\ell=(1+\frac{T}{4})\frac{H}{16}\frac{W}{16}$ used in
the main text.

\subsection{Self-Attention in the DiT}
\label{app:self-attn}
Let $d$ be the model width and $h$ the number of heads (with $d_h=d/h$).
For a sequence of length $\ell$:
\begin{align}
\text{Q,K,V projections:} \qquad &
3 \times 2\,\ell d^2 \;=\; 6\,\ell d^2 \nonumber\\
\text{Attention logits }(QK^\top): \qquad &
2\,\ell^2 d \nonumber\\
\text{Weighted sum }(AV): \qquad &
2\,\ell^2 d \nonumber\\
\text{Output projection:} \qquad &
2\,\ell d^2 \,.
\end{align}
Summing on all $N$ DiT layers yields
\begin{equation}
\label{eq:self-attn}
F_{\text{self}} \;=\; N\times(8\,\ell d^2 \;+\; 4\,\ell^2 d) \,.
\end{equation}
(The head count $h$ cancels out, since $h\cdot d_h=d$.)

\subsection{Cross-Attention (Video $\to$ Text)}
\label{app:cross-attn}
Let $m$ be the number of text tokens and $d$ the shared width.
Assuming no KV cache (K,V recomputed each denoising step as it is done in the current official implementation) and one
cross-attention block per DiT layer:
\begin{align}
\text{Query from video:} \qquad &
2\,\ell d^2 \nonumber\\
\text{Keys/values from text:} \qquad &
4\,m d^2 \quad (\text{K and V}) \nonumber\\
\text{Attention products:} \qquad &
2\,\ell m d \;+\; 2\,\ell m d \;=\; 4\,\ell m d \nonumber\\
\text{Output projection:} \qquad &
2\,\ell d^2 \,.
\end{align}
Hence over the N layers
\begin{equation}
\label{eq:cross-attn}
F_{\text{cross}} \;=\; N\times(4\,\ell d^2 \;+\; 4\,m d^2 \;+\; 4\,\ell m d) \,.
\end{equation}
\textit{With KV caching}, the $4md^2$ term becomes once-per-video while the
$4\ell m d$ products remain per step. \textit{With windowed or factorized
attention}, $\ell$ or $m$ may be replaced by the effective window size.

\subsection{Transformer MLP}
\label{app:mlp}
With expansion factor $f$ and sequence length $\ell$, a two-layer MLP
$d\!\to\!fd\!\to\!d$ costs over all DiT layers
\begin{equation}
\label{eq:mlp}
F_{\text{mlp}} \;=\; N \times 4 f\,\ell d^2 \,.
\end{equation}

\subsection{Stacking Across $S$ Steps, and CFG}
\label{app:stacking}
Let $g$ denote the number of conditional forward passes (CGF) per denoising step
($g=2$ under classifier-free guidance).
Combining~\eqref{eq:self-attn}–\eqref{eq:mlp}, the DiT cost is
\begin{equation}
\label{eq:dit-total}
F_{\text{DiT}}(T,H,W;S,N,d,f,m,g)
\;=\; g\,S\ \times\big( F_{\text{self}} + F_{\text{cross}} + F_{\text{mlp}} \big)\,,
\end{equation}
with $\ell$ given by~\eqref{eq:ell}.

\subsection{Text Encoder}
\label{app:text-enc}
For a $L_{\text{text}}$-layer encoder (e.g., T5/CLIP-like) with width
$d_{\text{text}}$, expansion $f_{\text{text}}$, and $m$ tokens:
\begin{align}
\text{Self-attn per layer:} \quad
& 8\,m d_{\text{text}}^2 + 4\,m^2 d_{\text{text}} \nonumber\\
\text{FFN per layer:} \quad
& 4 f_{\text{text}}\, m d_{\text{text}}^2 \,.
\end{align}
For $p_{\text{text}}$ forward passes per video (e.g., $p_{\text{text}}=2$
for conditional and unconditional prompts),
\begin{equation}
\label{eq:text-total}
F_{\text{text}} \;=\;
p_{\text{text}}\,L_{\text{text}}\,
\big( 8\,m d_{\text{text}}^2 + 4\,m^2 d_{\text{text}} + 4 f_{\text{text}}\, m d_{\text{text}}^2 \big).
\end{equation}
This term is once-per-video, independent of $S$.

\subsection{Timestep Embedding MLP}
\label{app:temb}
Mapping a scalar diffusion step to a $d$-dim vector and injecting it into each
block via a small MLP with hidden width $d_\tau$:
\begin{equation}
\label{eq:temb}
F_{\tau} \;=\; gS ( 2\,d_{\tau}\,d \;+\; 14\,d^2).
\end{equation}
\subsection{VAE: Convolutions and Middle Attention}
\label{app:vae}

We account for the VAE cost as the sum of (i) all convolutional layers along the decoder and 
(ii) a 2D self-attention ``middle'' block evaluated independently per time slice.

\paragraph{Convolutional layers.}
For a 3D convolution with kernel $(k_t^{(j)},k_h^{(j)},k_w^{(j)})$, channels
$C_{\mathrm{in}}^{(j)}\!\to\!C_{\mathrm{out}}^{(j)}$ and output size $T^{(j)}\times H^{(j)}\times W^{(j)}$,
the cost is
\begin{equation}
\label{eq:conv3d}
F_{\text{conv3d}}^{(j)} \;=\;
2\,k_t^{(j)} k_h^{(j)} k_w^{(j)} \, C_{\mathrm{in}}^{(j)} C_{\mathrm{out}}^{(j)} \, T^{(j)} H^{(j)} W^{(j)}\,.
\end{equation}
Summing over the decoder path gives
$F_{\text{VAE,conv}} = \sum_{j=1}^{N_{\text{dec,conv}}} F_{\text{conv3d}}^{(j)}$,
with concrete per-layer shapes provided in Table~\ref{tab:vae_layers_decoder}.
WAN-2.1 VAE include a 2D self-attention middle block evaluated independently on each time slice
($L_\ast=H_\ast W_\ast$, channel width $C_\ast$):
\begin{equation}
\label{eq:vae-middle-attn}
F_{\text{VAE,mid-attn}}
\;=\; T_\ast \big( 8\,C_\ast^2 L_\ast \;+\; 4\,L_\ast^2 C_\ast \big).
\end{equation}

\paragraph{Middle self-attention (2D, per time slice).}
Let $C_\ast$ be the channel width at the middle resolution, and 
$T_\ast,H_\ast,W_\ast$ the temporal/spatial sizes (thus $L_\ast = H_\ast W_\ast$ tokens per time slice).
Using the derivation in Appendix~\ref{app:self-attn}, the middle attention cost is
\begin{equation}
\label{eq:vae-middle-attn}
F_{\text{VAE,mid-attn}}
\;=\; T_\ast \big( 8\,C_\ast^2 L_\ast \;+\; 4\,L_\ast^2 C_\ast \big)\!,
\end{equation}
where the final $2C_\ast^2 L_\ast$ term arises from the output projection and is included in the $8C_\ast^2 L_\ast$ term above.

\paragraph{WAN2.1 decoder instantiation (values).}
In WAN2.1, the VAE decoder starts from a latent grid 
$(T_0,H_0,W_0) = \big(\lceil T/4\rceil,\, H/8,\, W/8\big)$ with $z{=}16$ channels. 
A causal $3{\times}3{\times}3$ convolution expands this to $384$ channels, 
followed by a ``middle'' block consisting of two residual $3{\times}3{\times}3$ convolutions 
and a 2D self-attention layer applied independently per time slice. 
The decoder then progressively upsamples: two \emph{temporal+spatial} upsamplings 
(doubling $T,H,W$ and halving channels), followed by one purely \emph{spatial} upsampling 
(doubling $H,W$ and halving channels). 
Residual blocks (three per stage) refine features at each resolution, 
and a final $3{\times}3{\times}3$ convolution produces the RGB output at $(T,H,W)$. 

Table~\ref{tab:vae_layers_decoder} summarizes the dominant operators for FLOP accounting. 
Applying Eq.~\eqref{eq:conv3d} across these layers yields $F_{\text{VAE,conv}}$, 
while Eq.~\eqref{eq:vae-middle-attn} gives the middle-attention cost.

\begin{table}[h]
\centering
\caption{VAE decoder: representative dominant operators for FLOP accounting (layer $j$). 
It mirrors the encoder; $z{=}16$, $C_\ast{=}384$, middle resolution $(\lceil T/4\rceil, H/8, W/8)$.}
\label{tab:vae_layers_decoder}
\begin{tabular}{c c c c c c c}
\toprule
Stage $j$ & Op type & Kernel $(k_t,k_h,k_w)$ & $C_{\mathrm{in}}^{(l)} \rightarrow C_{\mathrm{out}}^{(l)}$ 
& $T^{(l)}$ & $H^{(l)}$ & $W^{(l)}$ \\
\midrule
D0 & conv3d & $(3,3,3)$ & $z \rightarrow 384$ & $\lceil T/4\rceil$ & $H/8$ & $W/8$ \\
Middle (RBs) & conv3d & $(3,3,3)$ & $384 \rightarrow 384$ & $\lceil T/4\rceil$ & $H/8$ & $W/8$ \\
Middle (attn 2D) & attn-2D & -- & $384 \rightarrow 384$ & $\lceil T/4\rceil$ & $H/8$ & $W/8$ \\
D1 (RBs) & conv3d & $(3,3,3)$ & $384 \rightarrow 384$ & $\lceil T/4\rceil$ & $H/8$ & $W/8$ \\
Up (time) & conv3d (time) & $(3,1,1)$ & $384 \rightarrow 2\!\times\!384$ & $\lceil T/2\rceil$ & $H/8$ & $W/8$ \\
Up (space) & conv2d (space) & $(1,3,3)$ & $384 \rightarrow 192$ & $\lceil T/2\rceil$ & $H/4$ & $W/4$ \\
D2 (RBs) & conv3d & $(3,3,3)$ & $192 \rightarrow 384$ & $\lceil T/2\rceil$ & $H/4$ & $W/4$ \\
Up (time) & conv3d (time) & $(3,1,1)$ & $384 \rightarrow 2\!\times\!384$ & $T$ & $H/4$ & $W/4$ \\
Up (space) & conv2d (space) & $(1,3,3)$ & $384 \rightarrow 192$ & $T$ & $H/2$ & $W/2$ \\
D3 (RBs) & conv3d & $(3,3,3)$ & $192 \rightarrow 192$ & $T$ & $H/2$ & $W/2$ \\
Up (space) & conv2d (space) & $(1,3,3)$ & $192 \rightarrow 96$ & $T$ & $H$ & $W$ \\
Head & conv3d & $(3,3,3)$ & $96 \rightarrow 3$ & $T$ & $H$ & $W$ \\
\bottomrule
\end{tabular}
\end{table}

\subsection{Total FLOPs and Leading-Order Scaling}
\label{app:total}
We finally obtain
\begin{equation}
\label{eq:ftotal}
F_{\text{total}}(H,W,T,S) \;=\;
F_{\text{text}} \;+\; F_{\text{VAE,conv}} \;+\; F_{\text{VAE,mid-attn}}
\;+\; F_{\tau} \;+\; F_{\text{DiT}}\,,
\end{equation}
with components given by~\eqref{eq:text-total}, \eqref{eq:conv3d},
\eqref{eq:vae-middle-attn}, \eqref{eq:temb}, and \eqref{eq:dit-total}.
Since $\ell$ grows linearly with $H$, $W$, and $T$ (Eq.~\ref{eq:ell}),
the $\ell^2 d$ and $\ell m d$ terms in $F_{\text{DiT}}$ dominate for typical
settings ($\ell \gg m$), yielding quadratic growth in $H$, $W$, and $T$, and
linear growth in $S$.

\paragraph{Scope and caveats.}
(i) FlashAttention and fused kernels reduce memory traffic and constants but do
not change FLOP counts. (ii) KV caching changes only the cross-attention
$4md^2$ term from per-step to once-per-video. (iii) Windowed or factorized
attention replaces $\ell$ (or $m$) by an effective window size, altering
quadratic scaling. (iv) If activations or norms become bandwidth-bound, the
proportionality between FLOPs and latency weakens; our WAN2.1 measurements on
H100 indicated compute-bound behavior over the operating points considered.

\newpage
\section{Theoretical Compute-Bound Thresholds for DiT Blocks}
\label{sec:compute_bound_threshold}

We estimate the arithmetic intensity (FLOP per byte transferred between HBM and registers) for the main operations in DiT: the self-attention block (with FlashAttention) and the MLP. We then derive the compute-bound threshold $\ell^\star$ at which the operation’s intensity matches the hardware balance $\beta = \Theta_\text{peak} / B$.

Let $s$ be the byte size of a scalar (e.g., $s=2$ for BF16), and assume a fully optimized implementation that reads inputs and writes outputs only once from HBM, so each tensor contributes twice to memory traffic (read + write).

\paragraph{FlashAttention (forward).} We include only the matrix multiplications $QK^\top$ and $PV$ (not projections). The total FLOPs scale as $F_\text{attn} = 4 \ell^2 d$, and total memory transfer as $D_\text{attn} = 2 \ell d s$ (read inputs $Q,K,V$ and write output of size $\ell d$).

\[
\mathrm{AI}_\text{attn}(\ell) = \frac{F_\text{attn}}{D_\text{attn}} = \frac{4 \ell^2 d}{2 \ell d s} = \frac{2\ell}{s}
\quad\Rightarrow\quad
\ell^\star_\text{attn} = \frac{s \beta}{2}
\]

\paragraph{MLP block (GEMM)} The total FLOPs are $F_\text{mlp} = f \ell d^2$, and the memory transfer is $D_\text{mlp} = (f d^2 + \ell d + f \ell d )s$.

\[
\mathrm{AI}_\text{mlp}(\ell) = \frac{F_\text{mlp}}{D_\text{mlp}} = \frac{f \ell d}{(fd + \ell(1+f)) s}
\quad\Rightarrow\quad
\ell^\star_\text{mlp} = s \beta
\]

For $d=2048$, $s=2$, and $\beta = 295$ (H100 BF16), we find:

\[
\ell^\star_\text{attn} = \frac{2 \cdot 295}{2} = \mathbf{295}, \quad
\ell^\star_\text{mlp} = 2 \cdot 295= \mathbf{590}, \quad
\]

Thus, all MLP are compute-bound for $\ell > 590$, and attention becomes compute-bound for $\ell > 290$. In our WAN2.1 runs, $\ell \gg 10^4$, so both blocks operate far in the compute-bound regime.

\paragraph{Caveat.} These thresholds assume peak theoretical performance. In practice, we observe an empirical efficiency $\mu \approx 0.4$ for compute throughput on the H100. Similarly, the effective memory throughput often remains well below $B$ due to irregular access patterns and latency bottlenecks.

\paragraph{Other hardware.} Table~\ref{tab:flops_to_bandwidth} reports $\beta$ and the corresponding compute-bound thresholds for both attention and MLP blocks across a range of accelerators.

\begin{table}[h]
\centering
\caption{Approximate FLOP-to-bandwidth ratios ($\beta = \Theta_{\text{peak}}/B$) and corresponding compute-bound thresholds $\ell^\star$ for DiT blocks (BF16).}
\vspace{0.5em}
\begin{tabular}{lcccc}
\toprule
\textbf{Accelerator} & $\Theta_{\text{peak}}$ & $B$ & $\beta$ & $\ell^\star_\text{attn} \,/\, \ell^\star_\text{mlp}$ \\
& (TFLOP/s) & (TB/s) & (FLOP/byte) & \\
\midrule
NVIDIA H100 SXM & 989 & 3.35 & 295 & 295 / 590 \\
NVIDIA A100 SXM & 312 & 2.0 & 156 & 156 / 312 \\
RTX 4090 & 330 & 1.0 & 330 & 330 / 660 \\
NVIDIA L4 & 121 & 0.3 & 605 & 605 / 1210 \\
TPU v6 & 918 & 1.6 & 574 & 574 / 1148 \\
AMD M3250X & 2500 & 6.0 & 417 & 417 / 834 \\
Intel Gaudi3 & 1678 & 3.7 & 453 & 453 / 906 \\
\bottomrule
\end{tabular}
\label{tab:flops_to_bandwidth}
\end{table}

All realistic settings in WAN2.1 yield $\ell \gg 10^4$, even for low-resolution and short-duration inputs. Thus, both MLP and attention blocks operate well beyond the compute-bound threshold on all tested accelerators.

\end{document}